\pdfoutput=1
\documentclass[11pt]{article}

\usepackage[]{acl}

\usepackage{times}
\usepackage{latexsym}

\usepackage{graphicx}
\usepackage{amsmath}
\usepackage{amssymb}
\usepackage{amsfonts}
\usepackage{bbm}

\usepackage[linewidth=1pt]{mdframed}

\usepackage{xcolor}
\usepackage{amssymb}
\usepackage{algorithm}
\usepackage{bbding}
\usepackage{pifont}
\usepackage{wasysym}
\usepackage{multirow}
\usepackage{algpseudocode} 
\usepackage{booktabs}

\usepackage{amsthm,multicol,enumerate}
\usepackage{makecell}
\usepackage{graphicx}
\usepackage{color,xcolor}
\usepackage{subcaption}
\usepackage{enumitem}
\usepackage{soul}
\usepackage{array}

\usepackage{array}
\usepackage{multirow}
\usepackage{soul}
\usepackage{amsmath}
\usepackage{times}
\usepackage{latexsym}
\usepackage{graphicx}
\usepackage{amsfonts}
\usepackage{adjustbox}
\usepackage{tabularx}
\usepackage{etoolbox}
\usepackage{pgf}
\usepackage{etoc}
\usepackage[T1]{fontenc}
\usepackage[utf8]{inputenc}

\usepackage{microtype}

\usepackage{inconsolata}

\usepackage{soul}

\title{The Effect of Similarity Measures on Accurate Stability Estimates for Local Surrogate Models in Text-based Explainable AI}

\author{Christopher Burger \\
  Dept of Computer Science\\
  University of Mississippi\\
  \texttt{cburger@olemiss.edu} \\\And
  Charles Walter \\
  Dept of Computer Science\\
  University of Mississippi\\
  \texttt{cwwalter@olemiss.edu} \\\And
  Thai le \\
  Dept of Computer Science\\
  Indiana University\\
  \texttt{tle@iu.edu}
  }

\begin{document}
\maketitle
\begin{abstract}

%The popularity of recent black-box models has increased dramatically as their effectiveness continues to increase. However, this increase generally comes at the cost of added complexity. These complex black-box models lack justification for why a decision has been made, which has reduced their adoption in areas with serious consequences. One of the most popular remedies for this lack of explainability are local surrogate models, which attempt to provide explanation for a small set of inputs. But as local surrogates are themselves models, they are subject to weaknessess like any other model.
Recent work has investigated the vulnerability of local surrogate methods to adversarial perturbations on a machine learning (ML) model's inputs, where the explanation is manipulated while the meaning and structure of the original input remains similar under the complex model. Although weaknesses across many methods have been shown to exist, the reasons behind why remain little explored. Central to the concept of adversarial attacks on explainable AI (XAI) is the similarity measure used to calculate how one explanation differs from another. A poor choice of similarity measure can lead to erroneous conclusions on the efficacy of an XAI method. Too sensitive a measure results in exaggerated vulnerability, while too coarse understates its weakness. We investigate a variety of similarity measures designed for text-based ranked lists, including Kendall's Tau, Spearman's Footrule, and Rank-biased Overlap to determine how substantial changes in the type of measure or threshold of success affect the conclusions generated from common adversarial attack processes. Certain measures are found to be overly sensitive, resulting in erroneous estimates of stability.

%Local Surrogate models have increased in popularity for use in explaining complex black-box models for diverse types of data, including text, tabular, and image. Despite their continued use, questions about their stability still persist. Stability, a property where similar instances result in similar explanations, has been shown to be lacking in explanations generated for tabular and image data. Here we explore the stability of explanations generated on textual data, and find that their stability is also lacking. We compare an existing greedy search greedy search based method with that of a genetic algorithm in an attempt to increase the efficiency of an innately slow process. These efficiency increases were not realized, but initial results show a more effective selection of features to be be perturbed, resulting in more natural post-attack text.
\end{abstract}

\section{Introduction}

Continuous advancements in machine learning and AI have enabled the proliferation of powerful and complex models throughout much of our daily lives. %\st{These advanced models are used in fields as disparate as}
These advanced models are used in many disparate fields,
but disciplines with especially high consequences for failure, like medicine or finance, must be even more concerned with model error.

All models are subject to errors, but, when failure occurs, how is blame to be assigned? %Determining how to handle the consequences of complex model failure is now subject to intense debate from all sides of society, government, academic, private, and corporate\cite{}.
%https://ai.google/responsibility/responsible-ai-practices/
%https://learn.microsoft.com/en-us/azure/machine-learning/concept-responsible-ai?view=azureml-api-2
%https://www.whitehouse.gov/briefing-room/statements-releases/2023/05/04/fact-sheet-biden-harris-administration-announces-new-actions-to-promote-responsible-ai-innovation-that-protects-americans-rights-and-safety/
Complicating this assignment of blame is the type of failure that occurs. A model built for bird classification that misidentifies an image of a pine warbler as a yellow-throated vireo likely has no appreciable cost of failure, except possible embarrassment. A collision detection system that interprets a person standing in a crosswalk as only aberrant noise is an immeasurably greater failure.

% \begin{figure}[htb]
%   \centering
%   \includegraphics[width=\linewidth]{ExampleImage.png}
% %  \vspace{2.0cm}
%   %\centerline{(a) Result 1}\medskip
% %

% %
% % \vspace{-10pt}
% \caption{Original versus Perturbed Text for two documents subject to Adversarial Attack against the Explanatory Method. Both attacks passed the threshold of 50$\%$ dissimilarity, but the first is subject to more perturbations of a lower quality. The generating process for the first explanation is, in this case, more stable than that of the second.}
% \label{fig:explanationsExample}
% %
% % \vspace{-10pt}
% \end{figure}

But when any model fails, it is important that its stakeholders are informed of why such a failure occurs. With ever increasing model complexity, it is no longer feasible to directly interpret the actions of the model. As a response to this complexity, the discipline of explanatory AI (XAI) seeks to develop tools to allow both developers and users of models to understand why a model works the way it does. A common method in XAI is to generate an approximation for the complex model that is inherently interpretable. The output of these approximations, or surrogate models, is then used to understand why a model has made some prediction. Of the surrogate models, local surrogates are largely used due to their increased fidelity since local surrogates concern themselves only with explaining an individual prediction (or at most a small subset) and do not attempt to explain the model as whole. But, as surrogate models \textit{are} models, they can also be subject to the same issues as the models they attempt to explain. 

\subsection{XAI Stability}

One such property is stability or robustness, where an insignificant change in input should result in a correspondingly insignificant change in output \cite{alvarezmelis2018robustness}. This property is leveraged in adversarial attacks, where a model that is unstable can return obviously erroneous results, given small changes to the input. Local surrogate models have been shown to lack stability in many types of data, including image, tabular, and text \cite{slack2020fooling,ivankay2022fooling,xaifooler}. That is, the explanation of the complex model can be appreciably different given an insignificant change in the input. If an XAI model cannot produce consistent explanations across very similar documents when the output of the complex model itself remains similar, then the explanations from the XAI method become suspect. Suspect explanations cannot be trusted, and so the original complex model remains opaque and may be prevented from use due to legal or social implications despite its superior efficacy over other models.

\begin{table}[tb]%\[!htb\]
    %\begin{mdframed}
    \footnotesize

    \begin{subtable}[t]{.48\textwidth}
    %\caption*{}
    \raggedright
    \setlength\tabcolsep{7 pt}
        \begin{tabular}{clc|clc}
            %\hline \\
           \multicolumn{3}{c}{\textbf{Original Explanation}} & \multicolumn{3}{c}{\textbf{Perturbed Explanation}} \vspace{2pt}
            \\
             \multicolumn{2}{c}{\textbf{Feature}} & \multicolumn{1}{c}{\textbf{Weight}}&  \multicolumn{2}{c}{\textbf{Feature}} & \multicolumn{1}{c}{\textbf{Weight}}  \\
            \hline \\
            1 & heartburn & 1.77 & 1 & heartburn &  2.16 \\
                2 & eat & 0.59 & 2 & choking & 0.35 \\
                3 & vomit & 0.35 & 3 & puked & 0.34 \\
                4 & choking & 0.26 & 4 & eat & 0.33 \\
                5 & feel & 0.17 & 5 & like & 0.11 \\
                6 & lot & 0.15 & 6 & pain & 0.09 \\
                & ... & & &  ... & \\
                
            \hline \\
        \end{tabular}
\end{subtable}%

    % \begin{subtable}[t]{.24\textwidth}
    %             \caption*{\textbf{Original Explanation}}
    %     \raggedright
    %         \begin{tabular}{@{\extracolsep{1pt}}clc|}
    %             \hline \\
    %              & \multicolumn{1}{c}{\textbf{Feature}} & \multicolumn{1}{c}{\textbf{Weight}}  \\
    %             \hline \\
    %             1 & heartburn & 1.77 \\
    %             2 & eat & 0.59 \\
    %             3 & vomit & 0.35 \\
    %             4 & choking & 0.26 \\
    %             5 & feel & 0.17 \\
    %             6 & lot & 0.15 \\
    %             & ... & \\
    %             \hline \\
    %         \end{tabular}
    % \end{subtable}%
    %    \begin{subtable}[t]{.24\textwidth}
    %            \caption*{\textbf{Perturbed Explanation}}
    %     \raggedright
    %         \begin{tabular}{@{\extracolsep{1pt}}clc|}
    %             \hline \\
    %              & \multicolumn{1}{c}{\textbf{Feature}} & \multicolumn{1}{c}{\textbf{Weight}}  \\
    %             \hline \\
    %             1 & heartburn &  2.16 \\
    %             2 & choking & 0.35 \\
    %             3 & puked & 0.34 \\
    %             4 & eat & 0.33 \\
    %             5 & like & 0.11 \\
    %             6 & pain & 0.09 \\
    %              & ... & \\
    %             \hline \\
    %         \end{tabular}
    % \end{subtable}%
    %\bottomrule

    \begin{center}
     \textul{\textbf{Original Text} }  
    \end{center}
    
    I have a lot of heartburn and I feel like I'm choking when I eat. I also have a lot of stomach pain and I vomit a lot.
    \newline
    
    \begin{center}
     \textul{\textbf{Perturbed Text}}   
    \end{center}

    I have a \textbf{\textit{lots}} of heartburn and I feel like I'm choking when I eat. I also have a lot of stomach pain and I \textbf{\textit{puked}} a lot.

\begin{subtable}[t]{.48\textwidth}
        \caption*{\textbf{Comparative Similarities}}
        \raggedright
            \begin{tabular}{@{\extracolsep{4pt}}cccc}
                \toprule
                
                 %\multicolumn{1}{c}{\textbf{RBO}_{0.5}}
                    \multicolumn{1}{c}{\textbf{RBO}$_{0.7}$}
                    & \multicolumn{1}{c}{\textbf{RBO}$_{0.9}$}
                    & \multicolumn{1}{c}{\textbf{Jaccard}}
                    & \multicolumn{1}{c}{\textbf{Jaccard}$_w$}
                    \vspace{2pt}
                    \\
                    0.69 & 0.74 & 0.72 & 0.90 
                    \vspace{4pt}
                    \\
                    \multicolumn{1}{c}{\textbf{Kendall}}
                    & \multicolumn{1}{c}{\textbf{Kendall}$_w$}
                    & \multicolumn{1}{c}{\textbf{Spearman}}
                    & \multicolumn{1}{c}{\textbf{Spearman}$_w$}
                    \vspace{2pt}
                    \\
                    %\vspace{2pt}
                     0.10 & 0.48 & 0.50 & 0.50 \\
               \bottomrule
            \end{tabular}
    \end{subtable}%
    \caption{Perturbations generated under similarity measure \textbf{RBO$_{0.5}$} with final similarity of \textbf{0.75}  }
    \label{tab:comparative_perturbations_example}
    %\end{mdframed}
\end{table}

\subsection{Similarity Measures in XAI}
Previous work on XAI stability has focused on the \textit{existence} of instability given some fixed maximum amount of change applied to the input. This contrasts to standard adversarial attacks on (for example) classification models. These attacks often also have some perturbation limit, but the process has a clearly defined end point which is (if successful) a change in the predicted class. Previous work in XAI stability generally uses a similar approach to that of standard adversarial attacks, but lacks this clearly defined metric of success, and instead terminates at search exhaustion or perturbation limit \cite{xaifooler}. %The engine for selection of appropriate perturbations in the search process is its similarity measure, which controls the acceptance of perturbations and the overall determination of success or failure of the algorithm.  

\begin{table}[htb]%\[!htb\]
    
    \footnotesize
    \begin{subtable}[t]{.48\textwidth}
        %\caption*{}
        \raggedright
        \setlength\tabcolsep{3 pt}
            \begin{tabular}{clc|lc|lc}
                %\hline \\
               & \multicolumn{2}{c}{\textbf{Original}} & \multicolumn{2}{c}{\textbf{26.5\% Sim.}}&  \multicolumn{2}{c}{\textbf{28.9\% Sim.}} \vspace{2pt}
                \\
                 & \multicolumn{1}{c}{\textbf{Feature}} & \multicolumn{1}{c}{\textbf{Weight}}&  \multicolumn{1}{c}{\textbf{Feature}} & \multicolumn{1}{c}{\textbf{Weight}} & \multicolumn{1}{c}{\textbf{Feature}} & \multicolumn{1}{c}{\textbf{Weight}} \\
                \hline \\
                1 & dogs & 2.60 & tennis & 3.16 & tennis & 3.04 \\
                2 & balls & 2.10 & dogs & 2.65 & dogs & 2.68 \\
                3 & helpful & 0.44 & balls & 1.61 & balls & 2.22 \\
                4 & fetching & 0.39 & adore & 0.69 & love & 0.47 \\
                5 & love & 0.29 & useful & 0.49 & fetching & 0.42 \\
                6 & play & 0.07 & toy & 0.34 & toy & 0.39 \\
                7 & tennis & 0.04 & wish & 0.03 & helpful & 0.39 \\
                8 & wish & 0.03 & fetches & 0.02 & wish & 0.09 \\
                \hline \\
            \end{tabular}
    \end{subtable}%
    %    \begin{subtable}[t]{.24\textwidth}
    %     \caption*{\textbf{Perturbed Explanation}}
    %     \raggedright
    %         \begin{tabular}{@{\extracolsep{1pt}}ccc|}
    %             \hline \\
    %              & \multicolumn{1}{c}{\textbf{Feature}} & \multicolumn{1}{c}{\textbf{Weight}}  \\
    %             \hline \\
    %             1 & heartburn &  2.16 \\
    %             2 & choking & 0.35 \\
    %             3 & puked & 0.34 \\
    %             4 & eat & 0.33 \\
    %             5 & like & 0.11 \\
    %             6 & pain & 0.09 \\
    %              & ... & \\
    %             \hline \\
    %         \end{tabular}
    % \end{subtable}%
    %\bottomrule

        \begin{center}
     \textul{\textbf{Original Text}}   
    \end{center}
    
    i love dogs ! though i wish mine was more helpful while i play tennis . fetching balls . . .
    \newline

        \begin{center}
     \textul{\textbf{Perturbed Text - 26.5\% Similarity}}
    \end{center}

    i \textbf{\textit{adore}} dogs ! though i wish mine was more \textbf{\textit{useful}} while i \textbf{\textit{toy}} tennis . \textbf{\textit{fetches}} balls . . .
    \newline

        \begin{center}
     \textul{\textbf{Perturbed Text - 28.9\% Similarity}}
    \end{center}

    i love dogs ! though i wish mine was more helpful while i \textbf{\textit{toy}} tennis . fetching balls . . .
    \\

% \begin{subtable}[t]{.48\textwidth}
%         \caption*{\textbf{Comparative Similarities}}
%         \raggedright
%             \begin{tabular}{@{\extracolsep{4pt}}cccc}
%                 \toprule
%                  %\multicolumn{1}{c}{\textbf{RBO}_{0.5}}
%                     \multicolumn{1}{c}{\textbf{RBO}_{0.7}}
%                     & \multicolumn{1}{c}{\textbf{RBO}_{0.9}}
%                     & \multicolumn{1}{c}{\textbf{Jaccard}}
%                     & \multicolumn{1}{c}{\textbf{Jaccard}_w}
%                     \\
                    
%                     0.69 & 0.74 & 0.72 & 0.90 

%                     &\multicolumn{1}{c}{\textbf{Kendall}}
%                     & \multicolumn{1}{c}{\textbf{Kendall}_w}
%                     & \multicolumn{1}{c}{\textbf{Spearman}}
%                     & \multicolumn{1}{c}{\textbf{Spearman}_w}
%                     \\
                    
%                      0.10 & 0.48 & 0.50 & 0.50 \\
%                \bottomrule
%             \end{tabular}
%     \end{subtable}%
%\vspace{5pt}

\hrule
\caption{Perturbed explanations with close similarity values (calculated with respect to similarity measure RBO$_{0.5}$) despite substantial differences in perturbation rate.}
\label{tab:minimal_perturbations}
\end{table}

As the existence of instability has already been demonstrated, we seek to refine this knowledge asking \textit{ ``what is the importance of the similarity measure used to guide the adversarial search process?''}.
The similarity measure is the engine for the selection of appropriate perturbations in the search process and directly controls the acceptance of perturbations and the overall determination of the success or failure of the algorithm.  
In particular, we ask: Are certain similarity measures superior to others in providing quality adversarial explanations or accurate results in terms of XAI robustness? If so, under what conditions? Specifically, how do different measures compare in terms of \textit{sensitivity}, or propensity to show a difference in similarity with respect to another measure. In Table \ref{tab:comparative_perturbations_example} a perturbed explanation with 75\% similarity to the original calculated using the measure RBO$_{0.5}$ (Section \ref{sec:sim_measures}) is compared with eight other measures, resulting in up to 65\% lower similarity between documents only by choosing a new measure.

\begin{table*}[t]%\[!htb\]
    \footnotesize
    %\toprule
    %\vspace{5pt}

   \begin{center}
         % \multicolumn{12}{c}{\textbf{Original Text}} 
          \textbf{Original Text}
   \end{center}
   
    \vspace{5pt}

``I have been having headaches for a while now. They are usually on the left side of my head and are very painful. I also get nausea, vomiting, and sensitivity to light and sound. I have tried taking over-the-counter pain relievers, but they don't seem to help much.''
\\
    \begin{center}
         % \multicolumn{12}{c}{\textbf{Word Replacements Required to Satisfy Threshold}}  
         \textbf{Word Replacements Required to Satisfy Threshold}
    \end{center}

    \vspace{5pt}

    %     \begin{subtable}[h]{.2\textwidth}
    %     %\caption*{30\% Threshold}
    %     \setlength\tabcolsep{3 pt}
    %     \raggedright
    %         \begin{tabular}{llc}
    %            % \hline \\
    %              & \multicolumn{1}{c}{\textbf{Feature}} & \multicolumn{1}{c}{\textbf{Weight}}  \\
    %             \hline \\
    %             1 & headaches & 3.06 \\
    %             2 & sensitivity & 0.27 \\
    %             13 & \textit{assuming} & 0.22 \\
    %             3 & \textit{strived} & 0.26 \\
    %             10 & sound & 0.15 \\
    %             18 & vomiting & 0.04 \\
    %             & ... & 
    %         \end{tabular}
    % \end{subtable}% 
   \begin{subtable}[h]{.25\textwidth}
        \caption*{60\% Threshold}
        \raggedright
        \setlength\tabcolsep{2 pt}
            \begin{tabular}{lcc|}
                \hline \\
                 \multicolumn{1}{c}{\textbf{Original}}  & \multicolumn{1}{c}{\textbf{}} & \multicolumn{1}{c}{\textbf{Perturbed}}  \\
                \hline \\

                  pain & $\implies$ & agony \\
                 &  &  \\
                 &  &  \\
                 &  &  \\
                 &  &  \\
                 &  &  \\
                 & &
            \end{tabular}
    \end{subtable}%
       \begin{subtable}[h]{.25\textwidth}
        \caption*{50\% Threshold}
        \raggedright
        \setlength\tabcolsep{2 pt}
            \begin{tabular}{lcc|}
                \hline \\
                 \multicolumn{1}{c}{\textbf{Original}}  & \multicolumn{1}{c}{\textbf{}} & \multicolumn{1}{c}{\textbf{Perturbed}}  \\
                \hline \\

                  pain & $\implies$ & agony \\
                  help & $\implies$ & assistance \\
                 &  &  \\
                 &  &  \\
                 &  &  \\
                 &  &  \\
                 & &
            \end{tabular}
    \end{subtable}%
    \begin{subtable}[h]{.25\textwidth}
        \caption*{40\% Threshold}
        \raggedright
        \setlength\tabcolsep{2 pt}
            \begin{tabular}{lcc|}
                \hline \\
                 \multicolumn{1}{c}{\textbf{Original}}  & \multicolumn{1}{c}{\textbf{}} & \multicolumn{1}{c}{\textbf{Perturbed}}  \\
                \hline \\

                 pain & $\implies$ & agony \\
                 seem& $\implies$ &appears  \\
                 help& $\implies$ &aiding  \\
                 tried& $\implies$ &strived  \\
                 &  &  \\
                 &  &  \\
                 & &
            \end{tabular}
    \end{subtable}%
    \begin{subtable}[h]{.25\textwidth}
        \caption*{30\% Threshold}
        \raggedright
        \setlength\tabcolsep{2 pt}
            \begin{tabular}{lcc}
                \hline \\
                 \multicolumn{1}{c}{\textbf{Original}}  & \multicolumn{1}{c}{\textbf{}} & \multicolumn{1}{c}{\textbf{Perturbed}}  \\
                \hline \\

                 pain & $\implies$ & agony \\
                 seem& $\implies$ &appears  \\
                 help& $\implies$ &aiding  \\
                 tried& $\implies$ &strived  \\
                 having& $\implies$ & assuming  \\
                 side& $\implies$ &sides  \\
                 have&$\implies$ &ai
            \end{tabular}
    \end{subtable}%
    \hrule
    
\caption{Word Replacements required to satisfy each threshold. More stringent threshold generally necessitate more perturbations, which reduces overall textual quality.}
\label{tab:appendix_subset}
\end{table*}

\subsection{Minimal Perturbations}

Furthermore, does the choice of similarity measure affect the generation of \textit{minimum} viable perturbations? Where given a threshold of success, what are the fewest possible perturbations needed to reduce the similarity below this threshold while maintaining the structure and meaning of the original document. %This contrasts with standard adversarial attacks which have an obvious metric of success, a change to an erroneous prediction. But what determines a suitable difference between explanations is more subjective. However, once we have a threshold established we will be able to answer the question of is an XAI method stable with more clarity. 
Clearly, a method that consistently lacks stability when exposed to only a small number of perturbations is worse than one that requires a consistently large subset of the text to be changed. In Table \ref{tab:minimal_perturbations} we see that the first document (26.5\% similarity) requires substantially more perturbations to achieve a level of similarity comparable to that of the second attack (28.9\% similarity). The presence of a single perturbation that provides a significant decrease in similarity implies a much more serious instability compared to a method that requires multiple perturbations. Reducing the number of perturbations also helps maintain the quality of the perturbed text as ideal word replacements are difficult to generate given the vast search space. Table \ref{tab:appendix_subset} provides an example of the quality degradation process given repeated perturbation with the expanded text provided in Appendix \ref{apn:example}.

%\textit{minimal} viable perturbation that will sufficiently alter an explanation? To do so we need to establish some metric of success, just as the one that exists with classical adversarial perturbations. However, unlike a standard adversarial attack in which the alteration of the predicted class is the obvious metric of success, exactly what determines a suitable difference between explanations is more subjective. Once we have this threshold established we will be able to answer the question of is an XAI method stable with more clarity. A method that consistently lacks stability when exposed to only a small number of perturbations is clearly worse than one that that requires a consistently large subset of the text to be changed. In Fig. \ref{fig:explanationsExample} we see the first document requires more perturbations, and results in an appreciably less natural result than that of outcome of the second attack. The explanatory process that generated the first document is, in this instance, more stable than that of the second despite both explanations having similarity values that have fallen below some attack success criteria.

%In Fig. \ref{fig:explanationsExample} we see that the original text's semantic meaning is retained while there exists an appreciable difference between the explanations. But are these perturbations minimal? Can we achieve similar, or better, results with fewer perturbations? 
\subsection{Restriction to Natural Language}

We will restrict our focus to XAI models for text-based data. Our reasons are as follows: (1) Text-based models are ubiquitous, as natural language is a primary communication mechanism. (2) As natural language is woven into the fabric of our daily experiences, we as humans have an inherently robust sense for understanding the meaning of a document and so can judge the quality of a text document's explanation effectively without any formal procedure or training. This innate ability to judge the quality of an explanation allows us to choose reasonable thresholds, reducing some of what is an inherently subjective decision. (3) We can appreciably simplify the comparison process between similarity measures. Restricting ourselves to only a single data type avoids ambiguity when comparing results from a given similarity measure between types of data. Additionally, we can constrain the number of measures used, allowing use of those that may be more effective on text, but less transferable to other data. (4) Our depth of investigation can increase significantly. Testing XAI stability is a computationally intensive process. Thus, focusing on one data type allows a much broader range of thresholds and similarity measures to be tested given a fixed amount of time and computational resources.

\subsection{Contributions} Our contributions are summarized as follows.
\begin{enumerate}%[leftmargin=\dimexpr\parindent-0.2\labelwidth\relax,noitemsep]
    \item Exploring the effect on the quality of stability estimates using different similarity measures to guide the adversarial search process in text-based XAI. Moreover, identifying measures that are unsuitable for use due to excessive sensitivity resulting in exaggerated indications of instability.
    \item Whether the choice of similarity measure has an impact on the number of perturbations to determine a successful attack. This allows greater discernment between the XAI methods and their comparative levels of stability. 
\end{enumerate}

\section{Background \& Related Work}
% XAI and  \textsc{Lime} type models, Adversarial Attacks on  \textsc{Lime}, Methods of these attacks, Minimum viable perturbations, What these address

% Prior work on XAI stability has emphasized on evaluating models using tabular or image data across various interpretation methods, which often use small perturbations to the input data to generate appreciably different explanations~\cite{alvarezmelis2018robustness,InterpretationNN,alvarezmelis2018robustness}, or generate explanations that consist of arbitrary features~\cite{slack2020fooling}. \textcite{garreau2020explaining,garreau2022looking}, showed that key features can be omitted from the resulting explanations by changing parameters and that artifacts of the explanation generation process could produce misleading explanations. 

Previous work on XAI stability has focused on evaluating models using tabular or image data in various interpretation methods, which often use small perturbations of the input data to generate appreciably different explanations \cite{alvarezmelis2018robustness,InterpretationNN,alvarezmelis2018robustness}, or generate explanations consisting of arbitrary features \cite{slack2020fooling}. Garreau et al. showed that key features can be omitted from the resulting explanations by changing parameters and that artifacts of the explanation generation process could produce misleading explanations \cite{garreau2020explaining,garreau2022looking}. This was further extended to the analysis to text data \cite{mardaoui2021analysis} but only with respect to fidelity rather than stability of the surrogate models.

Our work here is restricted to the least explored domain, text. Previous work directly involving adversarial perturbations for XAI exists but has focused on determining the existence of such perturbations rather than establishing which components in the XAI method are most vulnerable \cite{sinha2021perturbing,ivankay2022fooling,xaifooler}. 
Other relevant work in text domain includes ~\cite{ivankay2022fooling}, which utilized a gradient-based approach but assumed white-box access to the target model; and \cite{sinha2021perturbing}, which generated adversarial attacks against black-box XAI methods. However, their experiment design may have led to an underestimation of stability as explored in \cite{xaifooler} which investigated the inherent instability of the method of choice's (\textsc{Lime}) sampling process for text data, and provided an alternate search strategy focused on the preferential perturbation of features deemed unimportant.

\subsection{XAI Method Selection}
While there are many XAI methods available, we narrow our choices by selecting three important criteria: proven usage in critical applications, a level of generalizability to other XAI methods, and the explanations generated are to satisfy certain attributes that constitute an effective explanation. That is, concision, order, and weight. \cite{xaifooler} From these criteria, we choose Local Interpretable Model-agnostic Explanations (\textsc{Lime})~\cite{LIME_Ribeiro} as our target explanatory algorithm. %which is a popular explanation method for both text, image and tabular data.
\textsc{Lime} is a commonly used and referenced tool in XAI frameworks, which has been integrated into critical ML applications such as finance~\cite{gramegna2021shap} and healthcare~\cite{kumarakulasinghe2020evaluating,fuhrman2022review}. To explain a prediction, \textsc{Lime} trains a shallow, inherently explainable surrogate model such as Logistic Regression on training examples that are synthesized within a neighborhood of an individual prediction. The resulting explanation is an ordered collection of features and their weights from this surrogate model that satisfies our requirements for a quality explanation outlined above. For text data, explanations generated by \textsc{Lime} have features that are individual words contained within the document to be explained, which can be easily understood even by non-specialists. We note that our method is applicable to any XAI method that returns an explanation in this format, satisfying our criteria of generalizability.

We note that since our goal is to investigate more precisely, the importance of the similarity measure used to guide the adversarial process \textsc{Lime} is not the focus of our investigation but used as a familiar standard due to its substantial base of previous work and common use. As the prior work considers only if the perturbed input's explanation is sufficiently different at the \textit{end} of a process, often consisting of many perturbations, we instead ask what would be reasonable similarity thresholds for stopping this process early?

\section{Problem Formulation}

%To find the minimal perturbation amounts we must first establish a search strategy that generates the perturbations, and then choose reasonable thresholds to allow the search to terminate.
As our goal is fundamentally similar to previous work in that we seek the discovery of perturbations that induce instability in XAI methods, our constraints when generating adversarial perturbations are generally equivalent to previous work that has established the existence of instability (in particular \textsc{Lime}. That is, the \textit{ meaning} of the input is retained, as well as an identical predicted class for the perturbed input in the original model. We use the general perturbation process outlined in \cite{xaifooler} with one addition. Although we retain a maximum amount of perturbations, we also use a threshold of similarity, $\tau$, which is used to signify a successful attack. %Where as long as available perturbations subject to the maximum limit exist and the best candidate similarity is above the threshold $\tau$ the search process continues.

\subsection{Search Procedure}
%With our objective function defined, 
The general search process is focused on the comparison of explanations and not on locating the appropriate perturbations. We use the greedy search procedure standard to previous work, where the indices of the words within the original document are ordered by importance (or lack thereof) to the model to be explained. The importance is calculated by removing the word at the $i^{th}$ index and determining the change in the predicted probability. These indices are sorted, filtered by the constraints, and then iterated through where the word at index $i$ has perturbations generated to replace it. These perturbations are the $n$ nearest neighbors in some embedding space where the final replacement is chosen by the largest decrease in similarity.

\subsection{Similarity  Measures}\label{sec:sim_measures}
We assume that the explanations generated are ranked lists, ordered by importance to the surrogate model (as is standard for \textsc{Lime} and common with other XAI methods). Then it is natural to choose measures for similarity or distance designed specifically for ranked lists. From the candidates, we select four popular measures that represent two main areas of comparison, set-based overlap, and dissonance between paired features. For the following definitions, let $A$ and $B$ be ranked lists composed of unique features.

For measures that use sets as a primary component, we choose the ubiquitous Jaccard index, and Rank-biased Overlap. The Jaccard index Eq. (\ref{eq:Jaccard_Definition}) is simply the ratio of the size of the intersection over the size of the union for two given sets. Intuitive and inherently bounded within $[0,1]$, the Jaccard index provides a useful baseline to compare against other measures. However, its simplicity makes it fairly coarse to determine similarity; for $Jaccard(A,B) = 0.50$, a similarity of $50\%$, this requires that at least $\frac{1}{3}$ of the words in B have been perturbed (to a word not already in $A$). This is beyond the capability of current XAI perturbation methods to maintain the context of the original text consistently. However, the inherent instability (the difference in identical documents with altered initialization parameters) within \textsc{Lime} can at times be substantial, enough to make the measure sensitive to what is ultimately ``natural variation ''. This sensitivity can be reduced by applying weights to the elements, an extension that we include for comparison. 
 \begin{equation}\label{eq:Jaccard_Definition}
     Jaccard(A,B) = \frac{|A \cap B|}{|A \cup B|}
 \end{equation}
 Rank-biased Overlap (RBO) \cite{RBO} is a summation of successively larger intersections, with each weighted by a term in a convergent series. This weighting scheme is controlled by a parameter $p \in (0,1)$ that can be adjusted to ascribe more or less weight to the top k features, a useful property for explanations, as concision is an important factor of explanability. Unbounding the surrogate model's number of features results in many features of approximately zero weight, which for regression models often chosen as the surrogate, this is effectively a judgment that the feature has no importance to the model being explained. Although RBO was formulated to take in lists of arbitrarily large length Eq. \ref{eq:RBO_Definition} displays the version restricted to a list of known size. Here, the size, $d = max(|A|,|B|)$. 
 \begin{equation}\label{eq:RBO_Definition}
 \begin{gathered}
        RBO(A,B,p,d) = \\ \frac{|A_{:d} \cap B_{:d}|}{d}  p^d +  \frac{(1-p)}{p}*\sum_{i=1}^{d}p^i  \frac{|A_{:i} \cap B_{:i}|}{i} 
     \end{gathered}
 \end{equation}

For the measures of paired features, we choose Kendall's Tau Rank Distance and Spearman's footrule. Both measures are in common use and are closely related, each a specific instance of a general correlation coefficient. Generally both measures return a distance rather than a similarity, but can be easily converted as both have a simple maximum bound described below.

Kendall's Tau counts the number of pairwise inversions between $A$ and $B$ ( Eq.\ref{eq:Kendall}) where $\mathbbm{1}[\cdot]$ is the indicator function.
\begin{equation}\label{eq:Kendall}
    \sum_{i=1}^{max(|A|,|B|)} \mathbbm{1}[A[i] \neq B[i]]
\end{equation}

The above formulation is an extension with the ability to handle lists that are not of equal size. For such lists if $| |A| - |B| | \neq 0$ this value is added to the summation in Eq. (\ref{eq:Kendall}). Clearly, the maximum possible amount of dissonant pairs is $max(|A|,|B|)$, which provides the denominator when converting Eq. (\ref{eq:Kendall}) to similarity. Other measures, especially those that calculate distance, may not have an upper bound. For our analysis, we restrict ourselves to measures that have a bounded maximum distance and can therefore be easily converted to a similarity value in $[0,1]$.

Spearman's footrule (Eq. \ref{eq:Spearman}) is the sum of the difference between the location $i$ of each feature $a \in \mathbf{A}$ and its corresponding location $j$ in $\mathbf{B}$. 

\begin{equation}\label{eq:Spearman}
  \sum_{a \in \mathbf{A}} | i - j |
\end{equation}

Spearman's footrule is effectively the $l_1$ distance applied to ranked lists. Similarly to Kendall's Tau, the footrule is by default not suitable for disjoint lists, but the footrule distance is bounded, with a maximum total distance of $\lfloor \frac{|\mathbf{A}|^2}{2}\rfloor$ with an individual element having at most $|\mathbf{A}| - 1$ possible distance. This leads to a natural choice for a penalty value for missing elements of $\frac{|\mathbf{A}|}{2}$, since for two completely disjoint lists we have $\sum_{i = 1}^{|\mathbf{A}|}\frac{|\mathbf{A}|}{2} = \frac{|\mathbf{A}|^2}{2}$. Using a penalty, the total maximum footrule distance can increase (Appendix \ref{apn:spearman}), but we note that this adjustment is not required as the proportion of disjoint elements between explanations is often small. As such, the original bound still remains useful even without the added penalty factor, which may avoid concerns of an arbitrary or ill-justified choice penalty. The weighted implementation for the footrule renders this concern even less meaningful, as the disjoint features within the explanation are concentrated within features of low importance and so are assigned very little weight. We note that the extension to disjoint lists can induce asymmetry depending on the construction of the measure. Our work calculates the adjusted footrule value dependent on the original explanation being compared to perturbed.

%Observations from the experiments conducted indicate that distances beyond the original bound are very rare, and the adjustment to total possible distance is often not required.

%(Talk about maximum distance, and the useful properties of penalty maxlength / 2)

Except for RBO which has inherent weighting, all of the previous measures have been seen in their unweighted format. The definitions are similar in structure, and we refer the reader to \cite{Sculley2007RankAF} \cite{GenDistKumar} for background and their derivations. For our experiments, we apply the same weights to each measure, which are the weights associated with the original unperturbed explanation, which are normalized with respect to the absolute value of the weight. For example, for characteristics $a, b$ with respective weights $0.25, 0.10$ then changing $a$ and $b$ equally under our measure (in terms of distance, dissonance, etc.) results in $a$ affecting the similarity more than $b$. These weighted measures are denoted using the subscript $_w$.

%(Mention the weighting strategy for each measure, and where the weights come from (normalized from the absolute value of the classifier weights with the exception of RBO with its user controlled weighting parameter, avoid explicitly using the formulas, instead cite the papers where the general formulation is derived)

% (Which similarity measures and what are reasonable thresholds to implement)

% (measures, two set based, two paired dissonance based, avoid measures not directly associated with ranked lists like $l_2$)
% \begin{enumerate}
%     \item Jaccard
%     \item Kendall
%     \item Spearman
%     \item Weighted versions (normalized explanation weights)? Likely too much computation here, also assigning the weights makes the resulting distance more difficult to interpret when normalized to [0,1]
%     \item RBO
% \end{enumerate}

\subsection{Success Thresholds}
We return to the question of where to stop the adversarial perturbation algorithm to balance fidelity with the original document and a sufficiently different explanation. As what constitutes a sufficiently different explanation is subjective, we instead select a range of values across each of the similarity measures to provide a broad comparison on common choices and to demonstrate the resulting quality of the perturbed document when a larger divergence from the original explanation is required. For our success threshold $\tau$, we choose levels of 30\%, 40\%, 50\%, and 60\% for the completion of the algorithm. Thresholds near 70\% begin to approach the inherent variation levels of \textsc{LIME} for certain datasets or nonstandard sampling rates \cite{xaifooler}. For thresholds lower than 30\%, the greedy search process is generally too inefficient to find appropriate perturbations for most similarity measures within the maximum perturbation limitations imposed. For measures that require a weighting parameter (\textbf{RBO}) we use the following values 0.5, 0.7, and 0.9. For this formulation, values approaching 0 increase the importance of the top features over the remaining features. Values close to 1.0 provide a less concentrated weight distribution. %which is counter-intuitive to the desired attributes of an effective explanation as then no feature becomes more important than any other. Instead of testing parameter values close to zero we have included the Jaccard index which ascribes no weight to any feature and is concerned only with a features presence in an explanation.

 \newcolumntype{Y}{>{\centering\arraybackslash}X}
\begin{table*}[h]
%\caption*{\textbf{Attack Success Rates}}
\footnotesize
\begin{tabularx}{\textwidth}{c *{11}{Y}}
\toprule
\multicolumn{1}{c}{\textbf{}}
& \multicolumn{1}{c}{\textbf{$\tau$}}
& \multicolumn{1}{c}{\textbf{RBO}$_{0.5}$}
& \multicolumn{1}{c}{\textbf{RBO}$_{0.7}$}
& \multicolumn{1}{c}{\textbf{RBO}$_{0.9}$}
& \multicolumn{1}{c}{\textbf{Jaccard}}
& \multicolumn{1}{c}{\textbf{Jaccard}$_w$}
& \multicolumn{1}{c}{\textbf{Kendall}}
& \multicolumn{1}{c}{\textbf{Kendall}$_w$}
& \multicolumn{1}{c}{\textbf{Spearman}}
& \multicolumn{1}{c}{\textbf{Spearman}$_w$}\\
%\cmidrule(lr){2-4} \cmidrule(l){5-7}
\cmidrule(lr){2-11}
& 30\%  & 0.12 &  0 &  0 & 0.02 & 0 & 0.95  & 0.52  & 0.14  & 0.02\\
\rotatebox[origin=c]{90}{\textbf{GB}}& 40\%  & 0.26 &  0.07 &  0 & 0.24 & 0 & 0.98  & 0.64  & 0.38  & 0.12\\
& 50\%  & 0.40 &  0.29 & 0.07 & 0.88 & 0 & 1  & 0.74  & 0.83  & 0.43\\
& 60\%  & 0.40 & 0.40 &  0.28 & 1 & 0.05 & 1  & 0.81  & 1  & 0.69\\
  \midrule

& 30\%  & 0.06 &  0.02 &  0 & 0.06 & 0 & 1  & 0.42  & 0.18  & 0.04\\
\rotatebox[origin=c]{90}{\textbf{S2D}}& 40\%  & 0.18 &  0.04 &  0 & 0.52 & 0 & 1  & 0.48  & 0.58  & 0.28\\
& 50\%  & 0.24 &  0.2 &  0.08 & 0.98 & 0.02 & 1  & 0.72  & 0.92  & 0.60\\
& 60\%  & 0.24 &  0.3 &  0.28 & 1 & 0.14 & 1  & 0.84  & 1  & 0.94
\\
\bottomrule 
\end{tabularx}
\caption{Attack Success Rates on \textsc{Lime} under DistilBERT.}
\label{tab:success_rates}
\end{table*}

%\vspace{-15pt}

\newcolumntype{Y}{>{\centering\arraybackslash}X}
\begin{table*}[h]
%\caption*{\textbf{Mean and Median Similarities}}
\footnotesize
\begin{tabularx}{\textwidth}{c *{20}{Y}}
\toprule
\multicolumn{1}{c}{\textbf{}}
& \multicolumn{1}{c}{\textbf{$\tau$}}
& \multicolumn{2}{c}{\textbf{RBO}$_{0.5}$}
& \multicolumn{2}{c}{\textbf{RBO}$_{0.7}$}
& \multicolumn{2}{c}{\textbf{RBO}$_{0.9}$}
& \multicolumn{2}{c}{\textbf{Jaccard}}
& \multicolumn{2}{c}{\textbf{Jaccard}$_w$}
& \multicolumn{2}{c}{\textbf{Kendall}}
& \multicolumn{2}{c}{\textbf{Kendall}$_w$}
& \multicolumn{2}{c}{\textbf{Spearman}}
& \multicolumn{2}{c}{\textbf{Spearman}$_w$}\\
%\cmidrule(lr){2-4} \cmidrule(l){5-7}
\cmidrule(lr){2-20}
& & $\mu$ & M & $\mu$ & M & $\mu$ & M & $\mu$ & M & $\mu$ & M & $\mu$ & M & $\mu$ & M & $\mu$ & M & $\mu$ & M \\ \addlinespace[3pt]
%\cmidrule(lr){3-20}
& 30\%  & 0.28 &  0.29 & - & - & - & - & 0.27  & 0.27   & - & - &  0.19 &  0.22 & 0.14 & 0.15 & 0.26  & 0.28  & 0.28  & 0.28\\
\rotatebox[origin=c]{90}{\textbf{GB}}& 40\%   &  0.34 & 0.32  & 0.38 & 0.39 & -  & - & 0.39 & 0.39   & - &  - &  0.26 & 0.29 & 0.19 & 0.24  & 0.35  & 0.36  & 0.37 & 0.38\\
& 50\%  & 0.40 &  0.43 &  0.48 & 0.48 & 0.46 & 0.47 & 0.47 & 0.46    & - & - &  0.32 &  0.32 & 0.28 & 0.33 & 0.47  & 0.48  & 0.47  & 0.47\\
& 60\%  & 0.40 &  0.43 &  0.52 & 0.53 & 0.57 & 0.58 & 0.55 & 0.55    & 0.59 & 0.59 &  0.37 &  0.40 & 0.33 & 0.36 & 0.56  & 0.56  & 0.54  & 0.55\\
\midrule

& 30\%  & 0.27 & 0.28  &  0.28 & 0.28 & - & - & 0.29 & 0.29   & - & - & 0.22 &  0.23 & 0.19 & 0.21 & 0.26  & 0.28  & 0.25  & 0.25\\
\rotatebox[origin=c]{90}{\textbf{S2D}}& 40\%  & 0.35 &  0.35 &  0.40 & 0.39 & - & -  & 0.37 & 0.38   & - & - &  0.29 &  0.29 & 0.28 & 0.33 & 0.37  & 0.38  & 0.37  & 0.37\\
& 50\%  & 0.42 &  0.46 &  0.48 & 0.49 & 0.49 & 0.49 & 0.47 & 0.48   & 0.47 & 0.47 &  0.36 &  0.38 & 0.39 & 0.44 & 0.47  & 0.48  & 0.45  & 0.46\\
& 60\%  & 0.42 &  0.46 &  0.55 & 0.58 & 0.58 & 0.58 & 0.56 & 0.57   & 0.56 & 0.57 & 0.38 & 0.39 & 0.49 & 0.51 & 0.56  & 0.57  & 0.57  & 0.58\\
\bottomrule
\end{tabularx}
\caption{Mean and Median Similarities of Successful Attacks.}
\label{tab:mean_median_similarities}
\end{table*}

\section{Experiment Setup}
To provide the raw material for our analysis, we generate batches of 50 adversarial examples using the algorithm in \cite{xaifooler}. Each batch is generated with respect to a similarity measure and a success threshold. Nine similarity measures are used: The Jaccard index, Kendall's Tau Rank Distance, and Spearman's footrule, each in their standard and weighted implementations and RBO, with weighting parameters 0.5, 0.7, and 0.9. Combined with the thresholds of 30\%, 40\%, 50\%, and 60\%, this results in 1,800 adversarial examples per data set. For the data sets, we used two of those included in \cite{xaifooler} and their associated pre-trained models. The first is the short length (average of 11 words) gender bias Twitter dataset (\textbf{GB}) \cite{dinan-etal-2020-multi} and the second being the moderate length (average of 29 words) symptoms to diagnosis dataset (\textbf{S2D}) (Kaggle). The final and longest-length dataset, IMDB movie reviews, proved computationally infeasible due to the excessive time requirements to generate the examples, approximately 125 days of continuous computation on a single A6000. The model to be explained is a DistilBERT \cite{sanh2019distilbert} fine-tuned on each respective dataset. Our interests concern the generalized properties of the similarity measures and not their specific performance associated with a given XAI method and base model, and so we choose for computational convenience a single efficient model to explain. %~\thai{justification on why we only run experiments on one model and if the results will generalize to other model/model types.}

%The chosen similarity measures guides the search process, and the success threshold is with respect to this measure. Statistics concerning the success rate of the attacks are presented in Table \ref{tab:success_rates}. Each completed attack, regardless of success or failure has its final explanation tested against every other similarity measure, the results of these comparisons are in Table \ref{tab:comparative_similarity_analysis}.

\newcolumntype{Y}{>{\centering\arraybackslash}X}
\begin{table*}[]
%\caption*{\textbf{Perturbed Word Rates}}
\footnotesize
\begin{tabularx}{\textwidth}{c *{11}{Y}}
\toprule
\multicolumn{1}{c}{\textbf{}}
& \multicolumn{1}{c}{\textbf{$\tau$}}
& \multicolumn{1}{c}{\textbf{RBO}$_{0.5}$}
& \multicolumn{1}{c}{\textbf{RBO}$_{0.7}$}
& \multicolumn{1}{c}{\textbf{RBO}$_{0.9}$}
& \multicolumn{1}{c}{\textbf{Jaccard}}
& \multicolumn{1}{c}{\textbf{Jaccard}$_w$}
& \multicolumn{1}{c}{\textbf{Kendall}}
& \multicolumn{1}{c}{\textbf{Kendall}$_w$}
& \multicolumn{1}{c}{\textbf{Spearman}}
& \multicolumn{1}{c}{\textbf{Spearman}$_w$}\\
%\cmidrule(lr){2-4} \cmidrule(l){5-7}
\cmidrule(lr){2-11}
& 30\%  & 0.14 & - &  - & 0.25 & - & 0.12  & 0.10  & 0.21  & 0.13\\
\rotatebox[origin=c]{90}{\textbf{GB}}& 40\%  & 0.14 & 0.19 &  - & 0.23 & - & 0.10  & 0.10  & 0.21  & 0.20\\
& 50\%  & 0.07 & 0.15 &  0.25 & 0.22 & - & 0.09  & 0.09  & 0.18  & 0.17\\
& 60\%  & 0.07 &  0.10 & 0.21 & 0.17 & 0.22 & 0.08  & 0.09  & 0.15  & 0.15\\
  \midrule

& 30\%  & 0.11 &  0.2 &  - & 0.23 & - & 0.06  & 0.08  & 0.17  & 0.16\\
\rotatebox[origin=c]{90}{\textbf{S2D}}& 40\%  & 0.12 &  0.12 & - & 0.20 & - & 0.05  & 0.07  & 0.17  & 0.15\\
& 50\%  & 0.05 &  0.15 &  0.19 & 0.17 & 0.21 & 0.04  & 0.07  & 0.14  & 0.15\\
& 60\%  & 0.05 &  0.09 &  0.17 & 0.13 & 0.20 & 0.04  & 0.05  & 0.10  & 0.13\\
\bottomrule
\end{tabularx}
\caption{Perturbed Word Rates for Successful Attacks}
\label{tab:perturbed_words}
\end{table*}

%\vspace{-100pt}

\newcolumntype{Y}{>{\centering\arraybackslash}X}
\begin{table*}[]
%\caption*{\textbf{Perturbed Document Quality}}
\footnotesize
\begin{tabularx}{\textwidth}{c *{20}{Y}}
\toprule
\multicolumn{1}{c}{\textbf{}}
& \multicolumn{1}{c}{\textbf{$\tau$}}
& \multicolumn{2}{c}{\textbf{RBO}$_{0.5}$}
& \multicolumn{2}{c}{\textbf{RBO}$_{0.7}$}
& \multicolumn{2}{c}{\textbf{RBO}$_{0.9}$}
& \multicolumn{2}{c}{\textbf{Jaccard}}
& \multicolumn{2}{c}{\textbf{Jaccard}$_w$}
& \multicolumn{2}{c}{\textbf{Kendall}}
& \multicolumn{2}{c}{\textbf{Kendall}$_w$}
& \multicolumn{2}{c}{\textbf{Spearman}}
& \multicolumn{2}{c}{\textbf{Spearman}$_w$}\\
%\cmidrule(lr){2-4} \cmidrule(l){5-7}
\cmidrule(lr){2-20}
& & \textbf{USE} & \textbf{PPL} & \textbf{USE} & \textbf{PPL} & \textbf{USE} & \textbf{PPL} & \textbf{USE} & \textbf{PPL} & \textbf{USE} & \textbf{PPL} & \textbf{USE} & \textbf{PPL} & \textbf{USE} & \textbf{PPL} & \textbf{USE} & \textbf{PPL} & \textbf{USE} & \textbf{PPL} \\ \addlinespace[3pt]
%\cmidrule(lr){3-20}
& 30\%  & 0.86 &  1.10 &  0 & 0 & 0 & 0  & 0.84  & 5.06  & 0 & 0 &  0.87 &  0.94 & 0.87 & 0.75 & 0.82  & 3.61  & 0.84  & 1.59\\
\rotatebox[origin=c]{90}{\textbf{GB}}& 40\%  & 0.85 &  2.04 &  0.84 & 2.34 & 0 & 0  & 0.82 & 3.76   & 0 & 0 & 0.88 & 0.68 & 0.88 & 0.66 & 0.82  & 1.79  & 0.81  & 1.77\\
& 50\%  & 0.89 & 0.51 & 0.84 & 1.98 & 0.80 & 5.66 & 0.83 & 2.25 & 0 & 0 & 0.89 &  0.65 & 0.88 & 0.61 & 0.84  & 1.88  & 0.84  & 1.52\\
& 60\%  & 0.89 &  0.51 &  0.87 & 0.89 & 0.81 & 2.12 & 0.85 & 1.60 & 0.82 & 4.51 &  0.90 & 0.45 & 0.89 & 0.50 & 0.85  & 1.26  & 0.85  & 1.55\\
\midrule
& 30\%  & 0.90 &  1.46 &  0.82 & 3.59 & 0 & 0  & 0.83 & 6.35  & 0 & 0 & 0.93 & 0.78 & 0.90 & 1.26 & 0.86  & 3.04  & 0.90  & 2.62\\
\rotatebox[origin=c]{90}{\textbf{S2D}}& 40\%  & 0.89 &  1.69 &  0.89 & 2.02 & 0 & 0 & 0.84 & 4.02  & 0 & 0 &  0.94 & 0.48 & 0.92 & 0.89 & 0.86  & 3.06  & 0.87  & 2.01\\
& 50\%  & 0.92 &  0.62 &  0.88 & 2.21 & 0.85 & 3.63  & 0.86 & 3.03  & 0.80 & 5.10 &  0.94 & 0.36 & 0.92 & 0.91 & 0.87  & 1.91  & 0.87  & 2.35\\
& 60\%  & 0.92 &  0.62 &  0.90 & 1.20 & 0.86 & 3.68  & 0.88 & 1.98  & 0.82 & 3.23 & 0.95 &  0.30 & 0.93 & 0.67 & 0.90  & 1.32  & 0.88  & 1.75\\
\bottomrule
\end{tabularx}
\caption{Perturbed Document Quality - \textbf{USE} is the Universal Sentence Encoder, \textbf{PPL} is perplexity.}
\label{tab:mean_semantic_similarities}
\end{table*}

\section{Results and Discussion}
All values in the following tables are calculated only for successful attacks. The symbol $\,$\textbf{-}$\,$ indicates non-applicability due to the absence of successful attacks on that particular combination of similarity measure, threshold, and data set. 

\noindent\textbf{Attack Success Rate} (Table \ref{tab:success_rates}): Immediately seen is the inappropriateness of certain similarity measures for use in text-based adversarial XAI attacks. Kendall's tau is extremely sensitive with almost 100\% attack success across every combination of threshold and data set. This level of sensitivity renders the measure useless here as the inherent instability of \textsc{Lime} is the cause, and little perturbation-induced instability is present. %This is not unexpected due to the definition of the measure and the inherent variation of \textsc{Lime} at the sampling rates chosen for practical use. Changes to the perturbed explanation as simple as a frameshift are strongly emphasized by the measure.   
This follows for the weighted version as well, though the success rates are not quite as high.
Jaccard, Spearman, and Spearman$_w$ also show excessive sensitivity for the higher similarity thresholds of 50\% and 60\%. 

Jaccard$_w$ and RBO$_{0.9}$ exhibit the opposite behavior, instead being coarse with few, if any, successful attacks under all but the most lenient threshold. This is not inherently negative, as both measures effectively require the top few characteristics to change, completely removing the explanation in the case of Jaccard$_w$ and at least a substantial decrease in the ranking for RBO$_{0.9}$. Since much of the weight of an explanation is often associated with the top features, these measures may prove useful as a fast heuristic for a substantial explanation difference.

The attack success rates are generally consistent between the datasets. As the sampling rate for \textsc{Lime} was optimized according to previous work, the inherent instability of the sampling process was kept similar despite the size difference.
\newline

\noindent \textbf{Perturbation Quality} (Tables \ref{tab:perturbed_words},\ref{tab:mean_semantic_similarities}):
The quality of the perturbed document follows the expected pattern of more perturbations resulting in lower quality. We exclude the Kendall measures from our discussion here as they are extremely sensitive and so few perturbations are required to meet the required threshold. Our quality measures are perplexity (\textbf{PPL}) and similarity calculated using cosine similarity with the universal sentence encoder (\textbf{USE}).

The coarseness of Jaccard$_w$ and RBO$_{0.9}$ results in the worst document quality with both the lowest similarity and the highest perplexity. Jaccard also has poor document quality. Both Spearman and Spearman$_w$ provide comparable results for the levels of perturbation given document quality. The RBO variants follow the expected trade-off of increased perturbation rates with subsequently reduced quality as the measure becomes coarser.
\newline

\vspace{-10pt}

\noindent \textbf{Minimal Perturbations}:
There are no strong indications for the choice in similarity measure being a major factor in finding minimal perturbations. We see no average ending similarities appreciably below the success threshold (with the exception of Kendall due to its sensitivity) (Table \ref{tab:mean_median_similarities}) and there are no notable signs of consistently low perturbation rates (Table \ref{tab:perturbed_words}). However, the search process is very greedy, taking the first viable perturbation that reduces the explanation similarity. An alternative search strategy would likely be superior when focused exclusively on minimal perturbations. Either with a threshold used to specify a minimum reduction in similarity in accepting a perturbation or with an entirely new algorithm that can explore the search space more effectively.  
\newline

%\vspace{-13pt}

\noindent \textbf{Overall}:
The Kendall and Jaccard measures are of limited use due to sensitivity or poor perturbed document quality. RBO in general provides a good balance between sensitivity and perturbed document quality across all thresholds at the cost of requiring manual adjustment of the weighting parameter for the given threshold. The Spearman measures show promise in comparison with RBO for more demanding thresholds but suffer from sensitivity for larger values. Spearman$_w$ may prove particularly useful if the original explanatory weights are deemed important or the fine-tuning of RBO too cumbersome. No particular measure stands out as substantially affecting the search procedure in terms of search efficiency or locating a set of minimal perturbations for a given threshold.

\section{Conclusion}
 We have shown that a poor choice in similarity measure can drastically skew the results of an adversarial attack, either by selecting a measure that is too sensitive and so overstating the weakness of the XAI method, or by choosing a measure too coarse and overstating the method's resilience. Practitioners and researchers should choose these measures judiciously, keeping in mind how sensitive a measure is to the inherent variation between explanations. Our findings confirm previous work in that measures that use both order and weight produce results with fewer perturbations and subsequently higher textual quality. However, no measure tested proved ideal in selecting for minimal perturbations. It appears necessary that the search process itself will also need adjustment as the current greedy approach is consistent regardless of the measure used. 

\section*{Limitations}

\textbf{Transferability to other XAI methods}: While the fundamentals of the discussion on similarity measures is not unique to \textsc{Lime}, our general conclusions on the fitness of particular similarity measures may not be applicable on different XAI methods. In particular, methods with less inherent instability may see more useful results from measures like Kendall's Tau, where its excessive sensitivity and subsequently exaggerated attack effectiveness may no longer pose a problem. 
\newline

\noindent\textbf{Computational Expenditure}: The computational resources needed to produce the adversarial examples were significant. Existing methods used to generate the attacks were not designed for multi-thousand collections of examples to be generated. Purpose-built tools focused on efficiency would allow larger numbers of samples to be reasonably generated, which may result in alternate conclusions.
\newline

\noindent\textbf{Other similarity measures}: The similarity measures used here were designed for the use of ranked lists. Other measures with this purpose or more generalized measures exist, and our conclusions may not be reflective of these untested measures.

\section*{Broader Impacts and Ethics Statement}
The authors anticipate there to be no reasonable
cause to believe use of this work would result in
harm, both direct or implicit. The authors disclaim
any conflicts of interest pertaining to this work.
\bibliography{custom,anthology}

\clearpage
\appendix
\onecolumn
\section{Additional Experimental Results}\label{apn:example}
\begin{table*}[!h]%\[!htb\]
    \caption{Perturbed Explanations at different attack success thresholds. Degradation in textual quality is apparent with continued perturbation. \textit{Italics} within an explanation indicate a word subject to perturbation or its perturbed form.}
    \footnotesize
    \begin{center}
          \textbf{Top-6 Features}
    \end{center}

    \begin{subtable}[h]{.2\textwidth}
        \caption*{Original}
        \raggedright
        \setlength\tabcolsep{3 pt}
            \begin{tabular}{clc|}
                \hline \\
                 & \multicolumn{1}{c}{\textbf{Feature}} & \multicolumn{1}{c}{\textbf{Weight}}  \\
                \hline \\
                1 & headaches & 2.80 \\
                2 & sensitivity & 0.30 \\
                3 & having & 0.26 \\
                4 & \textit{tried} & 0.25 \\
                5 & sound & 0.25 \\
                6 & vomiting & 0.21 \\
                & ... & 
                %\hline \\
            \end{tabular}
    \end{subtable}%
       \begin{subtable}[h]{.2\textwidth}
        \caption*{60\% Threshold}
        \raggedright
        \setlength\tabcolsep{3 pt}
            \begin{tabular}{clc|}
                \hline \\
                 & \multicolumn{1}{c}{\textbf{Feature}} & \multicolumn{1}{c}{\textbf{Weight}}  \\
                \hline \\
                % 1 & headaches & 3.10 \\
                % 5 & sensitivity & 0.19 \\
                % 6 & having & 0.16 \\
                % 9 & tried & 0.0.14 \\
                % 3 & sound & 0.25 \\
                %  8 & vomiting & 0.15 \\
                % & ... & 
                1 & headaches & 2.85 \\
                2 & counter & 0.30 \\
                3 & sound & 0.25 \\
                4 & don't & 0.23 \\
                5 & sensitivity & 0.19 \\
                6 & having & 0.16 \\
                & ...&
            \end{tabular}
    \end{subtable}%
        \begin{subtable}[h]{.2\textwidth}
        \caption*{50\% Threshold}
        \raggedright
        \setlength\tabcolsep{3 pt}
            \begin{tabular}{clc|}
                \hline\\
                 & \multicolumn{1}{c}{\textbf{Feature}} & \multicolumn{1}{c}{\textbf{Weight}}  \\
                \hline \\
                1 & headaches & 3.01 \\
                2 & having & 0.38 \\
                3 & sensitivity & 0.34 \\
                4 & taking & 0.24 \\
                5 & vomiting & 0.20 \\
                6 & don't &  0.18\\
                & ... & 
                % 1 & headaches & 3.01 \\
                % 3 & sensitivity & 0.34 \\
                % 2 & having & 0.38 \\
                % 12 & % tried & 0.26 \\
                % 13 & sound & 0.0.08 \\
                % 5 & vomiting & 0.20 \\
                % & ... & 
            \end{tabular}
    \end{subtable}%
        \begin{subtable}[h]{.2\textwidth}
        \caption*{40\% Threshold}
        \setlength\tabcolsep{3 pt}
        \raggedright
            \begin{tabular}{clc|}
                \hline\\
                 & \multicolumn{1}{c}{\textbf{Feature}} & \multicolumn{1}{c}{\textbf{Weight}}  \\
                \hline \\
                1 & headaches & 2.86 \\
                2 & sensitivity & 0.30 \\
                3 & having & 0.25 \\
                4 & \textit{strived} & 0.20 \\
                5 & counter & 0.17 \\
                6 & light & 0.17 \\
                & ... & \\
                % 1 & headaches & 3.06 \\
                % 2 & sensitivity & 0.27 \\
                % 13 & havinh & 0.25 \\
                % 3 & \textit{strived} & 0.20 \\
                % 16 & sound & 0.15 \\
                % 13 & vomiting & 0.10 \\
                % & ... & 
            \end{tabular}
    \end{subtable}%
        \begin{subtable}[h]{.2\textwidth}
        \caption*{30\% Threshold}
        \setlength\tabcolsep{3 pt}
        \raggedright
            \begin{tabular}{clc}
                \hline \\
                 & \multicolumn{1}{c}{\textbf{Feature}} & \multicolumn{1}{c}{\textbf{Weight}}  \\
                \hline \\
                1 & headaches & 3.06 \\
                2 & sensitivity & 0.27 \\
                3 & \textit{strived} & 0.26 \\
                4 & left & 0.25 \\
                5 & painful & 0.22 \\
                6 & counter & 0.21 \\
                & ... &
                % 1 & headaches & 3.06 \\
                % 2 & sensitivity & 0.27 \\
                % 13 & \textit{assuming} & 0.22 \\
                % 3 & \textit{strived} & 0.26 \\
                % 10 & sound & 0.15 \\
                % 18 & vomiting & 0.04 \\
                % & ... & 
            \end{tabular}
    \end{subtable}%
    \hrule
    \vspace{3pt}
    \begin{center}
         \textbf{New Locations for original Top-6 Features} 
    \end{center}
    \hrule
    %\cmidrule{1-12}
    
    \begin{subtable}[h]{.2\textwidth}
        %\caption*{Original}
        \raggedright
        \setlength\tabcolsep{3 pt}
            \begin{tabular}{llc|}
                %\hline \\
                 & \multicolumn{1}{c}{\textbf{Feature}} & \multicolumn{1}{c}{\textbf{Weight}}  \\
                \hline \\
                1 & headaches & 2.80 \\
                2 & sensitivity & 0.30 \\
                3 & \textit{having} & 0.26 \\
                4 & \textit{tried} & 0.25 \\
                5 & sound & 0.25 \\
                6 & vomiting & 0.21 \\
                & ... & 
                %\hline \\
            \end{tabular}
    \end{subtable}%
       \begin{subtable}[h]{.2\textwidth}
        %\caption*{60\% Threshold}
        \raggedright
        \setlength\tabcolsep{3 pt}
            \begin{tabular}{llc|}
                %\hline \\
                 & \multicolumn{1}{c}{\textbf{Feature}} & \multicolumn{1}{c}{\textbf{Weight}}  \\
                \hline \\
                1 & headaches & 3.10 \\
                5 & sensitivity & 0.19 \\
                6 & having & 0.16 \\
                9 & tried & 0.14 \\
                3 & sound & 0.25 \\
                8 & vomiting & 0.15 \\
                & ... & 

            \end{tabular}
    \end{subtable}%
        \begin{subtable}[h]{.2\textwidth}
        %\caption*{50\% Threshold}
        \raggedright
        \setlength\tabcolsep{3 pt}
            \begin{tabular}{llc|}
                %\hline\\
                 & \multicolumn{1}{c}{\textbf{Feature}} & \multicolumn{1}{c}{\textbf{Weight}}  \\
                \hline \\
                1 & headaches & 3.01 \\
                3 & sensitivity & 0.34 \\
                2 & having & 0.38 \\
                12 & tried & 0.26 \\
                13 & sound & 0.0.08 \\
                5 & vomiting & 0.20 \\
                & ... & 
            \end{tabular}
    \end{subtable}%
        \begin{subtable}[h]{.2\textwidth}
        %\caption*{40\% Threshold}
        \setlength\tabcolsep{3 pt}
        \raggedright
            \begin{tabular}{llc|}
               % \hline\\
                 & \multicolumn{1}{c}{\textbf{Feature}} & \multicolumn{1}{c}{\textbf{Weight}}  \\
                \hline \\
                1 & headaches & 3.06 \\
                2 & sensitivity & 0.27 \\
                13 & having & 0.25 \\
                3 & \textit{strived} & 0.20 \\
                16 & sound & 0.15 \\
                13 & vomiting & 0.10 \\
                & ... & 
            \end{tabular}
    \end{subtable}%
        \begin{subtable}[h]{.2\textwidth}
        %\caption*{30\% Threshold}
        \setlength\tabcolsep{3 pt}
        \raggedright
            \begin{tabular}{llc}
               % \hline \\
                 & \multicolumn{1}{c}{\textbf{Feature}} & \multicolumn{1}{c}{\textbf{Weight}}  \\
                \hline \\
                1 & headaches & 3.06 \\
                2 & sensitivity & 0.27 \\
                13 & \textit{assuming} & 0.22 \\
                3 & \textit{strived} & 0.26 \\
                10 & sound & 0.15 \\
                18 & vomiting & 0.04 \\
                & ... & 
            \end{tabular}
    \end{subtable}%

    \hrule
\vspace{5pt}
\begin{center}
\textbf{Original Text} 
\end{center}

I have been having headaches for a while now. They are usually on the left side of my head and are very painful. I also get nausea, vomiting, and sensitivity to light and sound. I have tried taking over-the-counter pain relievers, but they don't seem to help much.
\\

\begin{center}
\textbf{Perturbed - 60\% Similarity}    
\end{center}

I have been having headaches for a while now. They are usually on the left side of my head and are very painful. I also get nausea, vomiting, and sensitivity to light and sound. I have tried taking over-the-counter \textbf{\textit{agony}} relievers, but they don't seem to help much.
\\

\begin{center}
\textbf{Perturbed - 50\% Similarity}    
\end{center}

I have been having headaches for a while now. They are usually on the left side of my head and are very painful. I also get nausea, vomiting, and sensitivity to light and sound. I have tried taking over-the-counter \textbf{\textit{agony}} relievers, but they don't seem to \textbf{\textit{assistance}} much.
\\

\begin{center}
\textbf{Perturbed - 40\% Similarity}    
\end{center}

I have been having headaches for a while now. They are usually on the left side of my head and are very painful. I also get nausea, vomiting, and sensitivity to light and sound. I have \textbf{\textit{strived}} taking over-the-counter \textbf{\textit{agony}} relievers, but they don't \textbf{\textit{appears}} to \textbf{\textit{aiding}} much.
\\

\begin{center}
\textbf{Perturbed - 30\% Similarity}    
\end{center}

I have been \textbf{\textit{assuming}} headaches for a while now. They are usually on the left \textbf{\textit{sides}} of my head and are very painful. I also get nausea, vomiting, and sensitivity to light and sound. I \textbf{\textit{ai}} \textbf{\textit{strived}} taking over-the-counter \textbf{\textit{agony}} relievers, but they don't \textbf{\textit{appears}} to \textbf{\textit{aiding}} much.
\vspace{5pt}
%\bottomrule
\hrule
\end{table*}

\clearpage
\onecolumn
\section{Bounding Penalized Spearman's Footrule Maximum Distance}\label{apn:spearman}
Recall that Spearman's footrule possesses a maximum total distance (when the list is completely inverted) of $\lfloor \frac{|\mathbf{A}|^2}{2}\rfloor$ for $\mathbf{|A|} = \mathbf{|B|}, \; \mathbf{A} \cap \mathbf{B} = \mathbf{A}$ with an individual element having at most $|\mathbf{A}| - 1$ possible distance.
\newline

\noindent Given a penalty $p \geq 1$, for disjoint elements (i.e. there exists $e \in \mathbf{A}\; s.t \;e \notin \mathbf{B}$), we may have a total possible distance that exceeds the non-adjusted footrule.
\newline

\noindent Let $\mathbf{A}, \mathbf{B}$ be lists that are possibly disjoint and without loss of generality assume $\mathbf{|A|} \geq \mathbf{|B|} $. 
\newline

\noindent As the largest possible individual distance is $|\mathbf{A}| - 1$, assume $p \geq \mathbf{|A|} - 1$. Then trivially, the maximum distance is $p \mathbf{|A|}$.
\newline

\noindent Now assume $p < \mathbf{|A|} - 1$. The individual distances for a completely inverted list follow the patterns of:  $\mathbf{|A|} - 1, \mathbf{|A|} - 3, \mathbf{|A|} - 5 ... 1, 1, ... \mathbf{|A|} - 5, \mathbf{|A|} - 3, \mathbf{|A|} - 1$ for $\mathbf{|A|}$ even, and $\mathbf{|A|} - 1, \mathbf{|A|} - 3, \mathbf{|A|} - 5 ... 0 ... \mathbf{|A|} - 5, \mathbf{|A|} - 3, \mathbf{|A|} - 1$ for $\mathbf{|A|}$ odd.
\newline

\noindent We can view the new maximum penalty as the sum of two components: $p_g$, the individual distances greater or equal to the penalty, and $p_l$, those less than the penalty.
\newline

\noindent As $p < \mathbf{|A|} - 1$, there exists some distance $d \; s.t \; p \leq d \leq \mathbf{|A|} - 1$. Let this $d$ be located at the index $i$ within the pattern $\mathbf{|A|} - 1, \mathbf{|A|} - 3, \;...$ , where $\mathbf{|A|} - 1$ is the index $1$. The size of this subset of the above pattern is $\mathbf{|A|} - 2i$ and so its sum can be given as $\lfloor \frac{({|\mathbf{A}| - 2i})^2}{2}\rfloor$.
\newline

\noindent Then $p_g = \lfloor \frac{|\mathbf{A}|^2}{2}\rfloor - \lfloor \frac{({\mathbf{{|A|}} - 2i )}^2}{2}\rfloor$. And since $p \geq d$, then $p_l = p (\mathbf{{|A|}}-2i) -  \lfloor \frac{({|\mathbf{A}| - 2i})^2}{2}\rfloor$.
\newline 

\noindent So, the maximum distance for $p < \mathbf{|A|} - 1$  is given by $p_g + p_l =  
 \lfloor \frac{|\mathbf{A}|^2}{2}\rfloor + p (\mathbf{{|A|}} - 2i) -  2\lfloor \frac{({|\mathbf{A}| - 2i})^2}{2}\rfloor$.
\newline

% \noindent For $p = \frac{\mathbf{|A|}}{2}$ the new maximum distance is 

% \begin{equation*}
%  d =
%     \begin{cases}
%         1 & \text{if } \; \mathbf{|A|} \; even\\
%         0 & \text{if } \; \mathbf{|A|} \; odd
%     \end{cases}
% \end{equation*} 

%As $p < \mathbf{|A|} - 1$ then there exists some index $i \; s.t \; p_i < p \leq \mathbf{|A|} - 1$  

% Proof: 
% \newline

% For ordered list $|\mathbf{A}|$ consider the cases where $|\mathbf{A}|$ is odd and $|\mathbf{A}|$ is even. 
% \newline

% Odd:
% \newline

% For $|\mathbf{A}|$ odd the penalty $\frac{|\mathbf{A}|}{2}$ can be stated as $\frac{|\mathbf{A}|-1}{2} + \frac{1}{2}$. Where the penalty possesses an additional summand of 0.5 for each next odd size over that of the previous even list. 

% The total distance for $\mathbf{A}$ can be enumerated by the following pattern: $\mathbf{A}-1 + \mathbf{A}-3 + \mathbf{A}-5 + ... + \mathbf{A}-5 + \mathbf{A}-3 + \mathbf{A}-1 $ or $\sum_{i=0}^{|\mathbf{A}| / 2}|\mathbf{A}| - 2i - 1$

\end{document}